\documentclass{article}

\usepackage[final]{shakeshake}

\usepackage[utf8]{inputenc} 
\usepackage[T1]{fontenc}    
\usepackage{hyperref}       
\usepackage{url}            
\usepackage{booktabs}       
\usepackage{amsfonts}       
\usepackage{nicefrac}       
\usepackage{microtype}      
\usepackage[ruled, vlined, linesnumbered]{algorithm2e}
\usepackage[]{graphicx}
\usepackage{booktabs}
\usepackage{float}
\usepackage{caption}
\usepackage{subcaption}
\usepackage{color}
\usepackage[]{xcolor}
\usepackage{stmaryrd}
\usepackage{arydshln}

\makeatletter
\def\adl@drawiv#1#2#3{%
        \hskip.5\tabcolsep
        \xleaders#3{#2.5\@tempdimb #1{1}#2.5\@tempdimb}%
                #2\z@ plus1fil minus1fil\relax
        \hskip.5\tabcolsep}
\newcommand{\cdashlinelr}[1]{%
  \noalign{\vskip\aboverulesep
           \global\let\@dashdrawstore\adl@draw
           \global\let\adl@draw\adl@drawiv}
  \cdashline{#1}
  \noalign{\global\let\adl@draw\@dashdrawstore
           \vskip\belowrulesep}}
\makeatother

\definecolor{belize}{RGB}{41, 128, 185}

\title{Shake-Shake regularization}

\author{
  Xavier Gastaldi \\
  \texttt{xgastaldi.mba2011@london.edu} \\
}

\begin{document}

\maketitle

\begin{abstract}
The method introduced in this paper aims at helping deep learning practitioners faced with an overfit problem. The idea is to replace, in a multi-branch network, the standard summation of parallel branches with a stochastic affine combination.  Applied to 3-branch residual networks, shake-shake regularization improves on the best single shot published results on CIFAR-10 and CIFAR-100 by reaching test errors of 2.86\% and 15.85\%. Experiments on architectures without skip connections or Batch Normalization show encouraging results and open the door to a large set of applications. Code is available at \url{https://github.com/xgastaldi/shake-shake}.
\end{abstract}

\section{Introduction}

Deep residual nets \citep{He2015} were first introduced in the ILSVRC \& COCO 2015 competitions \citep{ILSVRC15,LinECCV14coco}, where they won the 1st places on the tasks of ImageNet detection, ImageNet localization, COCO detection, and COCO segmentation. Since then, significant effort has been put into trying to improve their performance. Scientists have investigated the impact of pushing depth \citep{He2016, Huang2016Densely}, width \citep{Zagoruyko2016WRN} and  cardinality \citep{xie2016groups, 45169, AbdiN16}. 

While residual networks are powerful models, they still overfit on small datasets. A large number of techniques have been proposed to tackle this problem, including weight decay \citep{doi:10.1162/neco.1992.4.4.473}, early stopping, and dropout \citep{JMLR:v15:srivastava14a}. While not directly presented as a regularization method, Batch Normalization \citep{IoffeS15} regularizes the network by computing statistics that fluctuate with each mini-batch. Similarly, Stochastic Gradient Descent (SGD) \citep{bottou-98x, Sutskever:2013:IIM:3042817.3043064} can also be interpreted as Gradient Descent using noisy gradients and the generalization performance of neural networks often depends on the size of the mini-batch (see \citet{KeskarMNST16}).

Pre-2015, most computer vision classification architectures used dropout to combat overfit but the introduction of Batch Normalization reduced its effectiveness (see \citet{IoffeS15, Zagoruyko2016WRN, HuangSLSW16}). Searching for other regularization methods, researchers started to look at the possibilities specifically offered by multi-branch networks. Some of them noticed that, given the right conditions, it was possible to randomly drop some of the information paths during training \citep{HuangSLSW16, larsson2016fractalnet}. 

Like these last 2 works, the method proposed in this document aims at improving the generalization ability of multi-branch networks by replacing the standard summation of parallel branches with a stochastic affine combination.

\subsection{Motivation}

Data augmentation techniques have traditionally been applied to input images only. However, for a computer, there is no real difference between an input image and an intermediate representation. As a consequence, it might be possible to apply data augmentation techniques to internal representations. Shake-Shake regularization was created as an attempt to produce this sort of effect by stochastically "blending" 2 viable tensors.

\subsection{Model description on 3-branch ResNets}

Let \(x_i\) denote the tensor of inputs into residual block \(i\). \(\mathcal{W}_{i}^{(1)}\) and \(\mathcal{W}_{i}^{(2)}\) are sets of weights associated with the 2 residual units. \(\mathcal{F}\) denotes the residual function, e.g. a stack of two 3x3 convolutional layers. \(x_{i+1}\) denotes the tensor of outputs from residual block \(i\).

A typical pre-activation ResNet with 2 residual branches would follow this equation:

\begin{equation} \label{eq:1}
x_{i+1} = x_i + \mathcal{F}(x_i,\mathcal{W}_{i}^{(1)}) + \mathcal{F}(x_i,\mathcal{W}_{i}^{(2)})
\end{equation}

Proposed modification: If \(\alpha_i\) is a random variable following a uniform distribution between 0 and 1, then during training:

\begin{equation} \label{eq:2}
x_{i+1} = x_i + \alpha_i\mathcal{F}(x_i,\mathcal{W}_{i}^{(1)}) + (1-\alpha_i)\mathcal{F}(x_i,\mathcal{W}_{i}^{(2)})
\end{equation}

Following the same logic as for dropout, all \(\alpha_i\) are set to the expected value of 0.5 at test time.

This method can be seen as a form of drop-path \citep{larsson2016fractalnet} where residual branches are scaled-down instead of being completely dropped (i.e. multiplied by 0). 

Replacing binary variables with enhancement or reduction coefficients is also explored in dropout variants like shakeout \citep{AAAI1611840} and whiteout \citep{whiteout}. However, where these methods perform an element-wise multiplication between an input tensor and a noise tensor, shake-shake regularization multiplies the whole image tensor with just one scalar \(\alpha_i\) (or \(1-\alpha_i\)).

\subsection{Training procedure}

\begin{figure}[h]
\begin{center}
\includegraphics[width=1\linewidth]{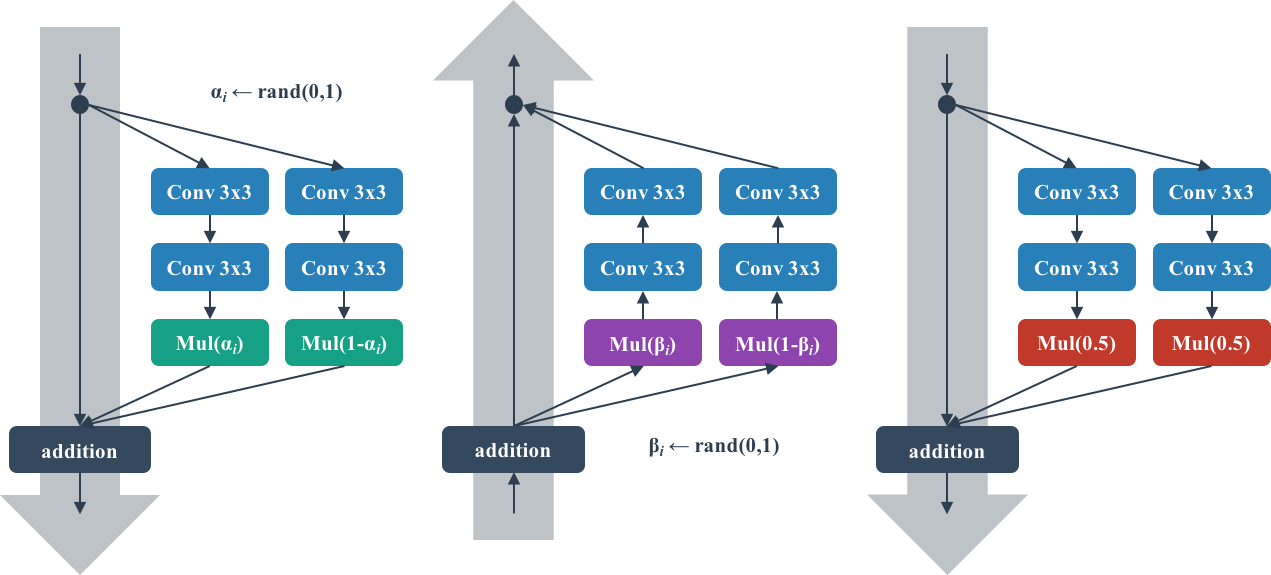}
\end{center}
\caption{\textbf{Left:} Forward training pass. \textbf{Center:} Backward training pass. \textbf{Right:} At test time.}
\label{modelimg}
\end{figure}

As shown in Figure~\ref{modelimg}, all scaling coefficients are overwritten with new random numbers before each forward pass. The key to making this work is to repeat this coefficient update operation before each backward pass. This results in a stochastic blend of forward and backward flows during training.

Related to this idea are the works of \citet{An:1996:EAN:1362174.1362187} and \citet{neelakantan2015adding}. These authors showed that adding noise to the gradient during training helps training and generalization of complicated neural networks. Shake-Shake regularization can be seen as an extension of this concept where gradient noise is replaced by a form of gradient augmentation.

\section{Improving on the best single shot published results on CIFAR}

\subsection{CIFAR-10}

\subsubsection{Implementation details}

The Shake-Shake code is based on {\tt fb.resnet.torch}\footnote{\url{https://github.com/facebook/fb.resnet.torch}} and is available at \url{https://github.com/xgastaldi/shake-shake}. The first layer is a 3x3 Conv with 16 filters, followed by 3 stages each having 4 residual blocks. The feature map size is 32, 16 and 8 for each stage. Width is doubled when downsampling. The network ends with a 8x8 average pooling and a fully connected layer (total 26 layers deep). Residual paths have the following structure: {\tt ReLU-Conv3x3-BN-ReLU-Conv3x3-BN-Mul}. The skip connections represent the identity function except during downsampling where a slightly customized structure consisting of 2 concatenated flows is used. Each of the 2 flows has the following components: 1x1 average pooling with step 2 followed by a 1x1 convolution. The input of one of the two flows is shifted by 1 pixel right and 1 pixel down to make the average pooling sample from a different position. The concatenation of the two flows doubles the width. Models were trained on the CIFAR-10 \citep{Krizhevsky09learningmultiple} 50k training set and evaluated on the 10k test set. Standard translation and flipping data augmentation is applied on the 32x32 input image. Due to the introduced stochasticity, all models were trained for 1800 epochs. Training starts with a learning rate of 0.2 and is annealed using a Cosine function \emph{without} restart (see \citet{loshchilov2016sgdr}). All models were trained on 2 GPUs with a mini-batch size of 128. Other implementation details are as in {\tt fb.resnet.torch}.

\subsubsection{Influence of Forward and Backward training procedures}

The base network is a 26 2x32d ResNet (i.e. the network has a depth of 26, 2 residual branches and the first residual block has a width of 32). "Shake" means that all scaling coefficients are overwritten with new random numbers before the pass. "Even" means that all scaling coefficients are set to 0.5 before the pass. "Keep" means that we keep, for the backward pass, the scaling coefficients used during the forward pass. "Batch" means that, for each residual block \(i\), we apply the same scaling coefficient for all the images in the mini-batch. "Image" means that, for each residual block \(i\), we apply a \emph{different} scaling coefficient for \emph{each} image in the mini-batch (see Image level update procedure below). 

\paragraph{Image level update procedure:} Let \(x_{0}\) denote the original input mini-batch tensor of dimensions 128x3x32x32. The first dimension « stacks » 128 images of dimensions 3x32x32. Inside the second stage of a 26 2x32d model, this tensor is transformed into a mini-batch tensor \(x_{i}\) of dimensions 128x64x16x16. Applying Shake-Shake regularization at the Image level means slicing this tensor along the first dimension and, for each of the 128 slices, multiplying the \(j^{th}\) slice (of dimensions 64x16x16) with a scalar \(\alpha_{i.j}\) (or \(1-\alpha_{i.j}\)).

\begin{table}[h]
\caption{Error rates (\%) on CIFAR-10. Results that surpass all competing methods by more than 0.1\% are \textbf{bold} and the overall best result is \textbf{\textcolor{belize}{blue}}.}

\label{results1}
\begin{center}
\begin{tabular}{cccccc}
\toprule
\multicolumn{3}{c}{} &\multicolumn{3}{c}{Model} \\
\cmidrule(lr){4-6}
\multicolumn{1}{c}{Forward} &\multicolumn{1}{c}{Backward} &\multicolumn{1}{c}{Level}  &\multicolumn{1}{c}{26 2x32d} &\multicolumn{1}{c}{26 2x64d} &\multicolumn{1}{c}{26 2x96d} \\
\midrule
Even	&Even	&n/a	&4.27	&3.76	&3.58\\
\cdashlinelr{1-6}
Even	&Shake	&Batch	&4.44	&-	&-\\
Shake	&Keep	&Batch	&4.11	&-	&-\\
Shake	&Even	&Batch	&3.47	&\textbf{3.30}	&-\\
Shake	&Shake	&Batch	&3.67	&\textbf{3.07}	&-\\
\cdashlinelr{1-6}
Even	&Shake	&Image	&4.11	&-	&-\\
Shake	&Keep	&Image	&4.09	&-	&-\\
Shake	&Even	&Image	&3.47	&\textbf{3.20}	&-\\
Shake	&Shake	&Image 	&3.55	&\textbf{2.98}	&\textbf{\textcolor{belize}{2.86}}\\
\bottomrule
\end{tabular}
\end{center}
\end{table}

\begin{figure}[h]
    \centering
    \begin{subfigure}[b]{0.45\textwidth}
        \includegraphics[width=\textwidth]{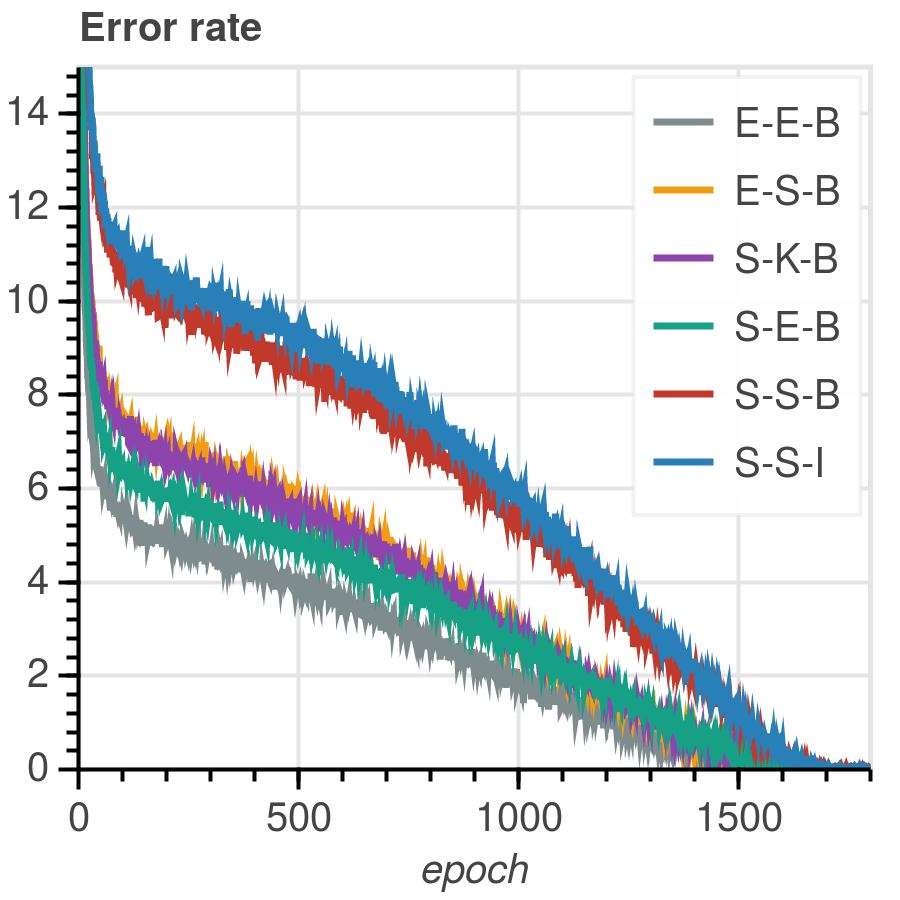}
    \end{subfigure}
    \begin{subfigure}[b]{0.45\textwidth}
        \includegraphics[width=\textwidth]{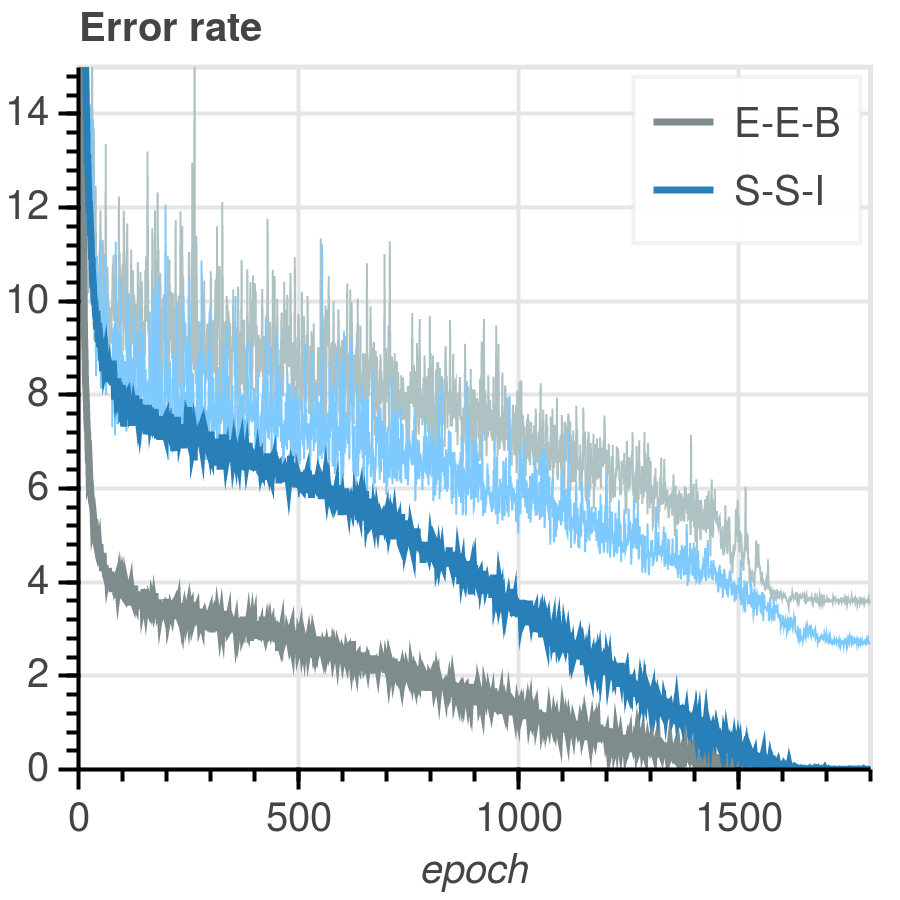}
    \end{subfigure}
    \caption{\textbf{Left:} Training curves of a selection of 32d models. \textbf{Right:} Training curves (dark) and test curves (light) of the 96d models.}\label{fig:training_curves}
\end{figure}

The numbers in Table~\ref{results1} represent the average of 3 runs except for the 96d models which were run 5 times. What can be observed in Table~\ref{results1} and Figure~\ref{fig:training_curves} is that "Shake-Keep" or "S-K" models (i.e. "Shake" $\shortrightarrow$ Forward  $\shortrightarrow$  "Keep" $\shortrightarrow$ Backward) do not have a particularly strong effect on the error rate. The network seems to be able to see through the perturbations when the weight update is done with the same ratios as during the forward pass. "Even-Shake" only works when applied at the "Image" level. "Shake-Even" and "Shake-Shake" models all produce strong results at 32d but the better training curves of "Shake-Shake" models start to make a difference when the number of filters of the first residual block is increased to 64d. Applying coefficients at the "Image" level seems to improve regularization.

\subsection{CIFAR-100}

The network architecture chosen for CIFAR-100 is a ResNeXt without pre-activation (this model gives slightly better results on CIFAR-100 than the model used for CIFAR-10). Hyperparameters are the same as in \citet{xie2016groups} except for the learning rate which is annealed using a Cosine function and the number of epochs which is increased to 1800. The network in Table~\ref{results2} is a ResNeXt-29 2x4x64d (2 residual branches with 4 grouped convolutions, each with 64 channels). Due to the combination of the larger model (34.4M parameters) and the long training time, fewer tests were performed than on CIFAR-10. 

\begin{table}[h]
\caption{Error rates (\%) on CIFAR-100. Results that surpass all competing methods by more than 0.5\% are \textbf{bold} and the overall best result is \textbf{\textcolor{belize}{blue}}.}
\label{results2}
\begin{center}
\begin{tabular}{ccccc}
\toprule
\multicolumn{4}{c}{} &\multicolumn{1}{c}{Model} \\
\cmidrule(lr){5-5}
\multicolumn{1}{c}{Forward} &\multicolumn{1}{c}{Backward} &\multicolumn{1}{c}{Level} &\multicolumn{1}{c}{Runs}  &\multicolumn{1}{c}{29 2x4x64d}\\
\midrule
Even	&Even	&n/a    &2	&\textbf{16.34}\\
\cdashlinelr{1-5}
Shake	&Even	&Image  &3	&\textbf{\textcolor{belize}{15.85}}\\
Shake	&Shake	&Image  &1 	&\textbf{15.97}\\
\bottomrule
\end{tabular}
\end{center}
\end{table}

Interestingly, a key hyperparameter on CIFAR-100 is the batch size which, compared to CIFAR-10, has to be reduced from 128 to 32 if using 2 GPUs.\footnote{As per notes in \url{https://github.com/facebookresearch/ResNeXt}} Without this reduction, the E-E-B network does not produce competitive results. As shown in Table~\ref{results2}, the increased regularization produced by the smaller batch size impacts the training procedure selection and makes S-E-I a slightly better choice.

\subsection{Comparisons with state-of-the-art results}

At the time of writing, the best single shot model on CIFAR-10 is a DenseNet-BC k=40 (3.46\% error rate) with 25.6M parameters. The second best model is a ResNeXt-29, 16x64d (3.58\% error rate) with 68.1M parameters. A small 26 2x32d "Shake-Even-Image" model with 2.9M parameters obtains approximately the same error rate. This is roughly 9 times less parameters than the DenseNet model and 23 times less parameters than the ResNeXt model. A 26 2x96d "Shake-Shake-Image" ResNet with 26.2M parameters, reaches a test error of 2.86\% (Average of 5 runs - Median 2.87\%, Min = 2.72\%, Max = 2.95\%).

On CIFAR-100, a few hyperparameter modifications of a standard ResNeXt-29 8x64d (batchsize, no pre-activation, longer training time and cosine annealing) lead to a test error of 16.34\%. Adding shake-even regularization reduces the test error to 15.85\% (Average of 3 runs - Median 15.85\%, Min = 15.66\%, Max = 16.04\%).

\begin{table}[h]
\caption{Test error (\%) and model size on CIFAR. Best results are \textbf{\textcolor{belize}{blue}}.}
\label{results3}
\begin{center}
\begin{tabular}{cccccc}
\toprule
\multicolumn{1}{c}{Method} &\multicolumn{1}{c}{Depth} &\multicolumn{1}{c}{Params} &\multicolumn{1}{c}{C10}  &\multicolumn{1}{c}{C100}\\
\midrule
Wide ResNet		&28	&36.5M    &3.8	&18.3\\
\cdashlinelr{1-5}
ResNeXt-29, 16x64d	&29	&68.1M    &3.58	  &17.31\\
\cdashlinelr{1-5}
DenseNet-BC (k=40)	&190	&25.6M    &3.46	 &17.18\\
\cdashlinelr{1-5}
C10 Model S-S-I	  	&26	&26.2M    &\textbf{\textcolor{belize}{2.86}}	&-\\
C100 Model S-E-I	&29	&34.4M    &-	&\textbf{\textcolor{belize}{15.85}}\\
\bottomrule
\end{tabular}
\end{center}
\end{table}

\section{Correlation between residual branches}

To check whether the correlation between the 2 residual branches is increased or decreased by the regularization, the following test was performed:

For each residual block: 
\begin{enumerate}  
\item Forward a mini-batch tensor \(x_{i}\) through the residual branch 1 (ReLU-Conv3x3-BN-ReLU-Conv3x3-BN-Mul(0.5)) and store the output tensor in \(y_{i}^{(1)}\). Do the same for residual branch 2 and store the output in \(y_{i}^{(2)}\).
\item Flatten these 2 tensors into vectors \(flat_{i}^{(1)}\) and \(flat_{i}^{(2)}\). Calculate the covariance between each corresponding item in the 2 vectors using an online version of the covariance algorithm.
\item Calculate the variances of \(flat_{i}^{(1)}\) and \(flat_{i}^{(2)}\) using an online variance algorithm.
\item Repeat until all the images in the test set have been forwarded. Use the resulting covariance and variances to calculate the correlation.
\end{enumerate}

\begin{figure}
    \centering
    \begin{subfigure}[b]{0.7\textwidth}
        \includegraphics[width=\textwidth]{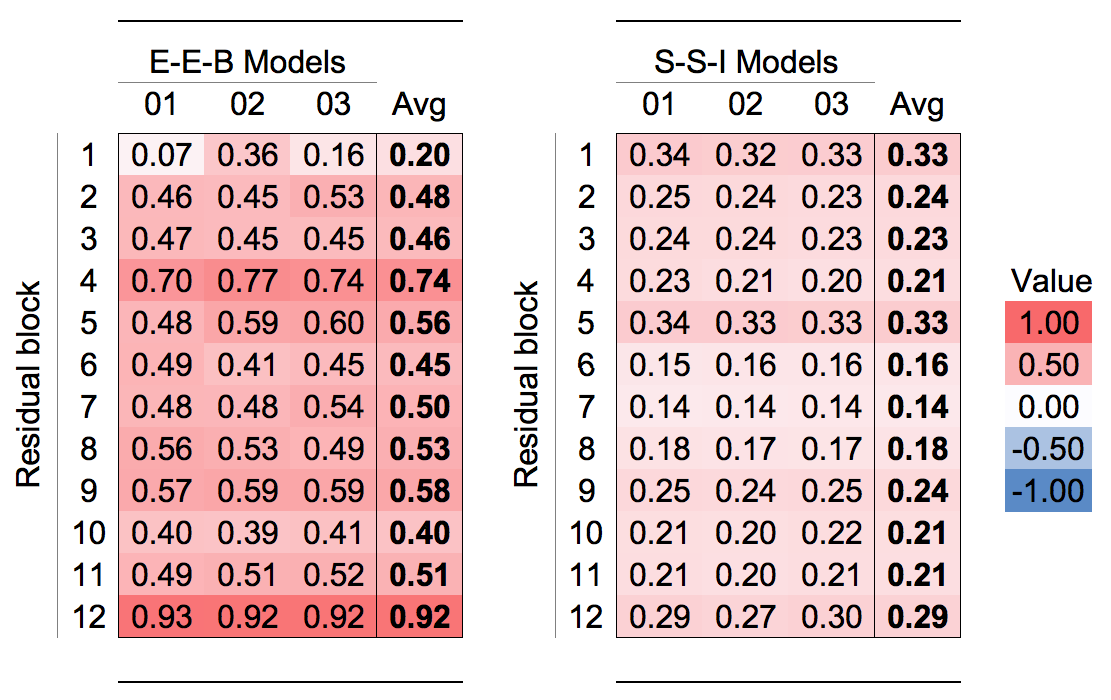}
    \end{subfigure}
    \caption{Correlation results on E-E-B and S-S-I models.}\label{fig:correlation1}
\end{figure}

This algorithm was run on CIFAR-10 for 3 EEB models and 3 S-S-I models both 26 2x32d. The results are presented in Figure~\ref{fig:correlation1}. The correlation between the output tensors of the 2 residual branches seems to be reduced by the regularization. This would support the assumption that the regularization forces the branches to learn something different.

One problem to be mindful of is the issue of alignment (see \citet{li_2016_ICLR}). The method above assumes that the summation at the end of the residual blocks forces an alignment of the layers on the left and right residual branches. This can be verified by calculating the layer wise correlation for each configuration of the first 3 layers of each block.

\begin{figure}
    \centering
    \begin{subfigure}[b]{1\textwidth}
        \includegraphics[width=\textwidth]{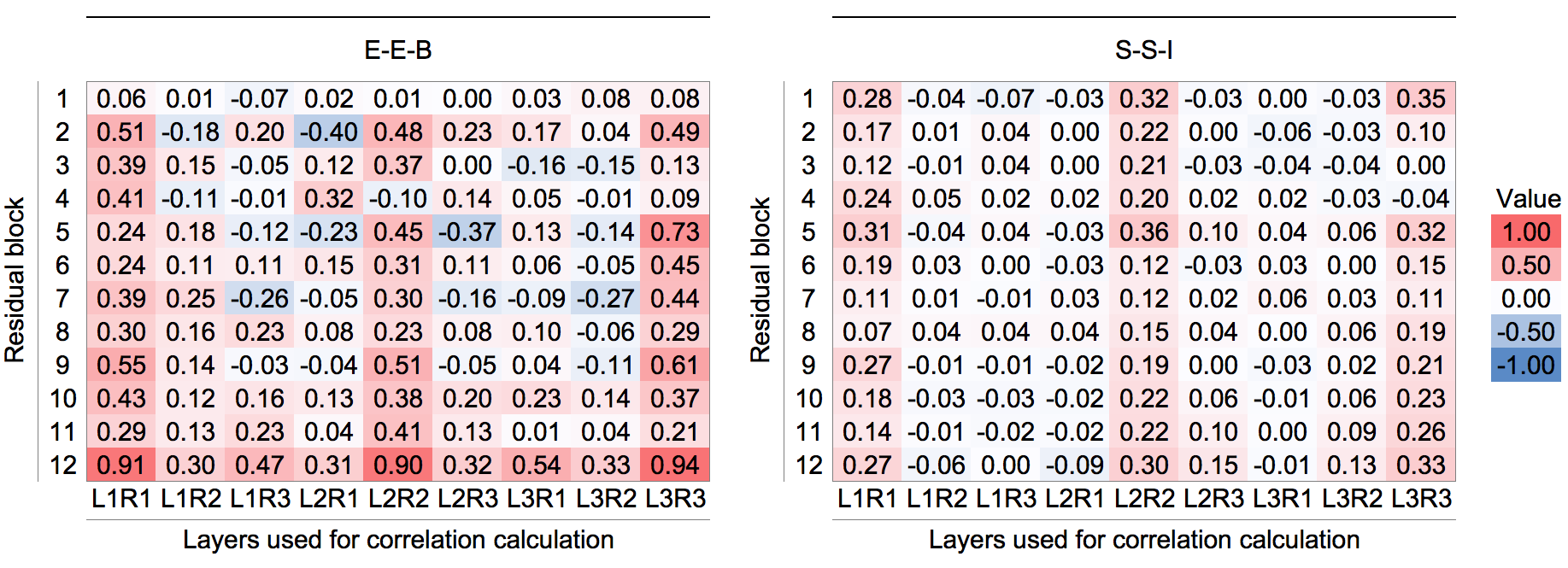}
    \end{subfigure}
    \caption{Layer-wise correlation between the first 3 layers of each residual block.}\label{fig:correlation2}
\end{figure}

The results are presented in Figure~\ref{fig:correlation2}. L1R3 for residual block \(i\) means the correlation between the activations of the first layer in \(y_{i}^{(1)}\) (left branch) and the third layer in \(y_{i}^{(2)}\) (right branch). Figure~\ref{fig:correlation2} shows that the correlation between the same layers on the left and right branches (i.e. L1R1, L2R2, etc..) is higher than in the other configurations, which is consistent with the assumption that the summation forces alignment.

\section{Regularization strength}

This section looks at what would happen if we give, during the backward pass, a large weight to a branch that received a small weight in the forward pass (and vice-versa).

Let \(\alpha_{i.j}\) be the coefficient used during the forward pass for image \(j\) in residual block \(i\). Let \(\beta_{i.j}\) be the coefficient used during the backward pass for the same image at the same position in the network. 

\begin{figure}[h]
    \centering
    \begin{subfigure}[b]{0.45\textwidth}
        \includegraphics[width=\textwidth]{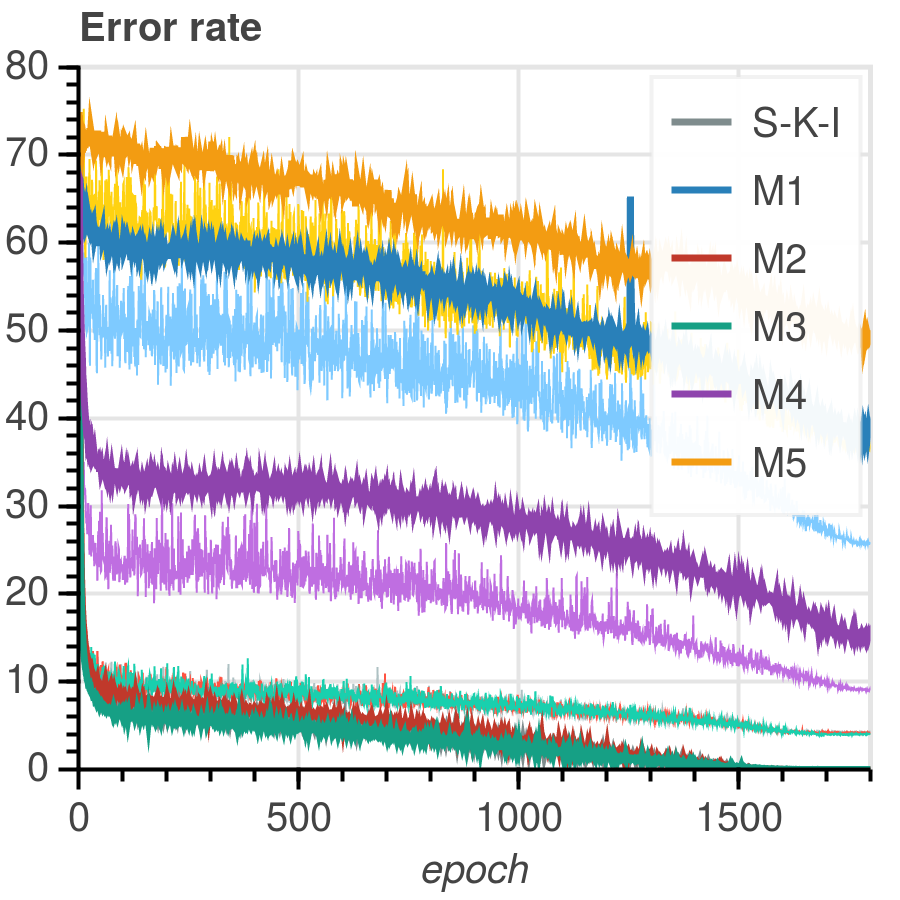}
    \end{subfigure}
    \begin{subfigure}[b]{0.38\textwidth}
        \includegraphics[width=\textwidth]{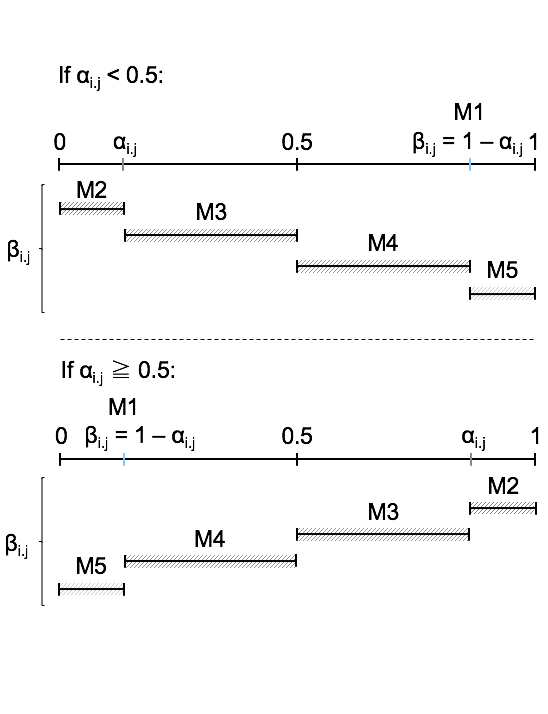}
    \end{subfigure}
    \caption{\textbf{Left:} Training curves (dark) and test curves (light) of models M1 to M5. \textbf{Right:} Illustration of the different methods in Table 4.}\label{fig:illustration}
\end{figure}

The first test (method 1) is to set \(\beta_{i.j}\) = 1 - \(\alpha_{i.j}\). All the tests in this section were performed on CIFAR-10 using 26 2x32d models at the Image level. These models are compared to a 26 2x32d Shake-Keep-Image model. The results of M1 can be seen on the left part of Figure~\ref{fig:illustration} (blue curve). The effect is quite drastic and the training error stays really high. 

Tests M2 to M5 in Table~\ref{methods} were designed to understand why Method 1 (M1) has such a strong effect. The right part of Figure~\ref{fig:illustration} illustrates Table~\ref{methods} graphically.

\begin{table}[h]
\caption{Update rules for \(\beta_{i.j}\).}
\label{methods}
\begin{center}
\begin{tabular}{ccc}
\toprule
\multicolumn{1}{c}{Method} &\multicolumn{1}{c}{\(\alpha_{i.j} < 0.5\)} &\multicolumn{1}{c}{\(\alpha_{i.j} \geq 0.5\)}\\
\midrule
S-S-I	&\(rand(0,1)\)	&\(rand(0,1)\)\\
S-E-I	&\(0.5\)	&\(0.5\)\\
M1		&\(1 - \alpha_{i.j}\)	&\(1 - \alpha_{i.j}\)\\
M2	&\(rand(0,1)*\alpha_{i.j}\)	&\(rand(0,1)*(1-\alpha_{i.j}) + \alpha_{i.j}\)\\
M3	&\(rand(0,1)*(0.5-\alpha_{i.j}) + \alpha_{i.j}\)	&\(rand(0,1)*(\alpha_{i.j}-0.5) + 0.5\)\\
M4	  	&\(rand(0,1)*(0.5-\alpha_{i.j}) + 0.5\)	&\(rand(0,1)*(0.5 - (1-\alpha_{i.j})) + (1 - \alpha_{i.j})\)\\
M5	&\(rand(0,1)*\alpha_{i.j} + (1-\alpha_{i.j})\)	&\(rand(0,1)*(1-\alpha_{i.j})\)\\
\bottomrule
\end{tabular}
\end{center}
\end{table}

What can be seen is that:

\begin{enumerate}  
\item The regularization effect seems to be linked to the relative position of \(\beta_{i.j}\) compared to \(\alpha_{i.j}\)
\item The further away \(\beta_{i.j}\) is from \(\alpha_{i.j}\), the stronger the regularization effect
\item There seems to be a jump in strength when 0.5 is crossed
\end{enumerate}

These insights could be useful when trying to control with more accuracy the strength of the regularization.

\section{Removing skip connections / Removing Batch Normalization}

One interesting question is whether the skip connection plays a role. A lot of deep learning systems don't use ResNets and making this type of regularization work without skip connections could extend the number of potential applications. 

\begin{figure}
    \centering
    \begin{subfigure}[b]{0.32\textwidth}
        \includegraphics[width=\textwidth]{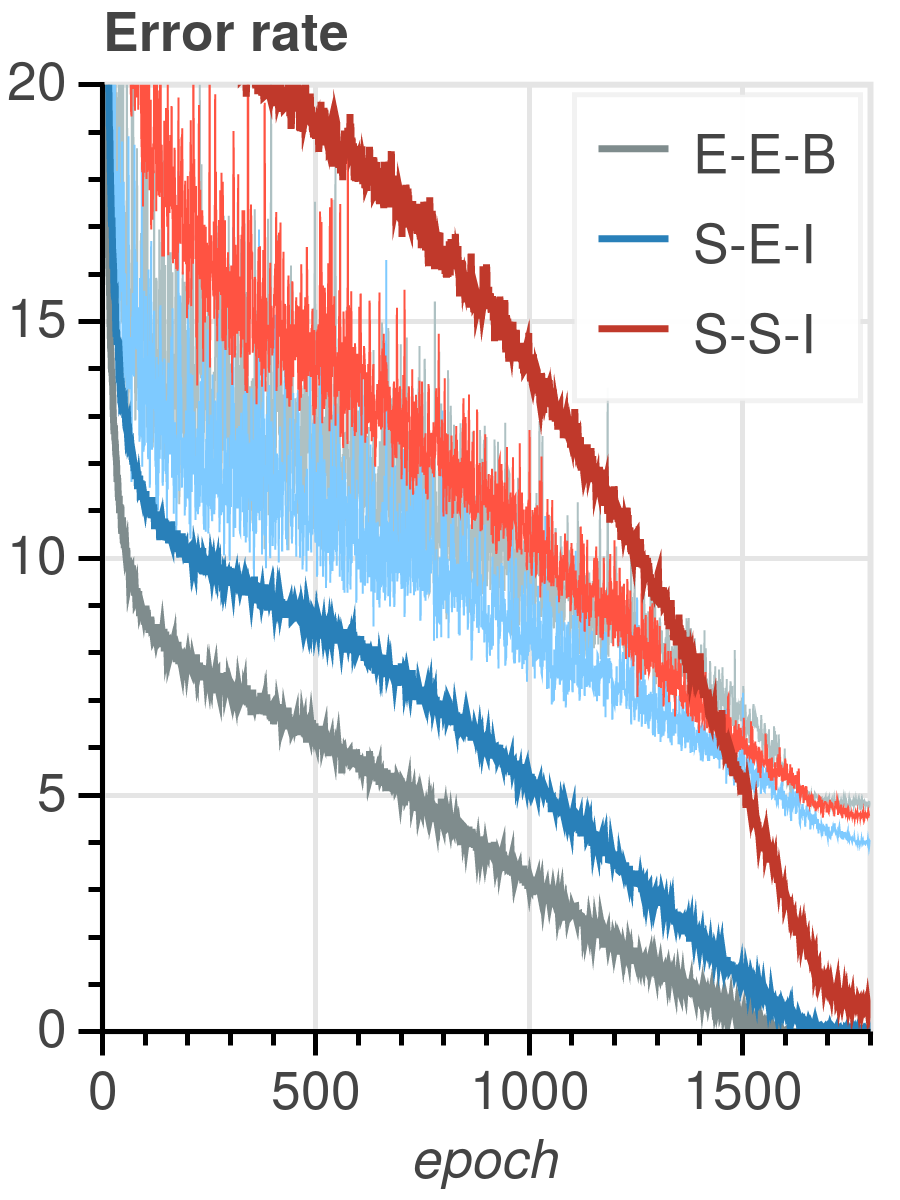}
    \end{subfigure}
    \begin{subfigure}[b]{0.32\textwidth}
        \includegraphics[width=\textwidth]{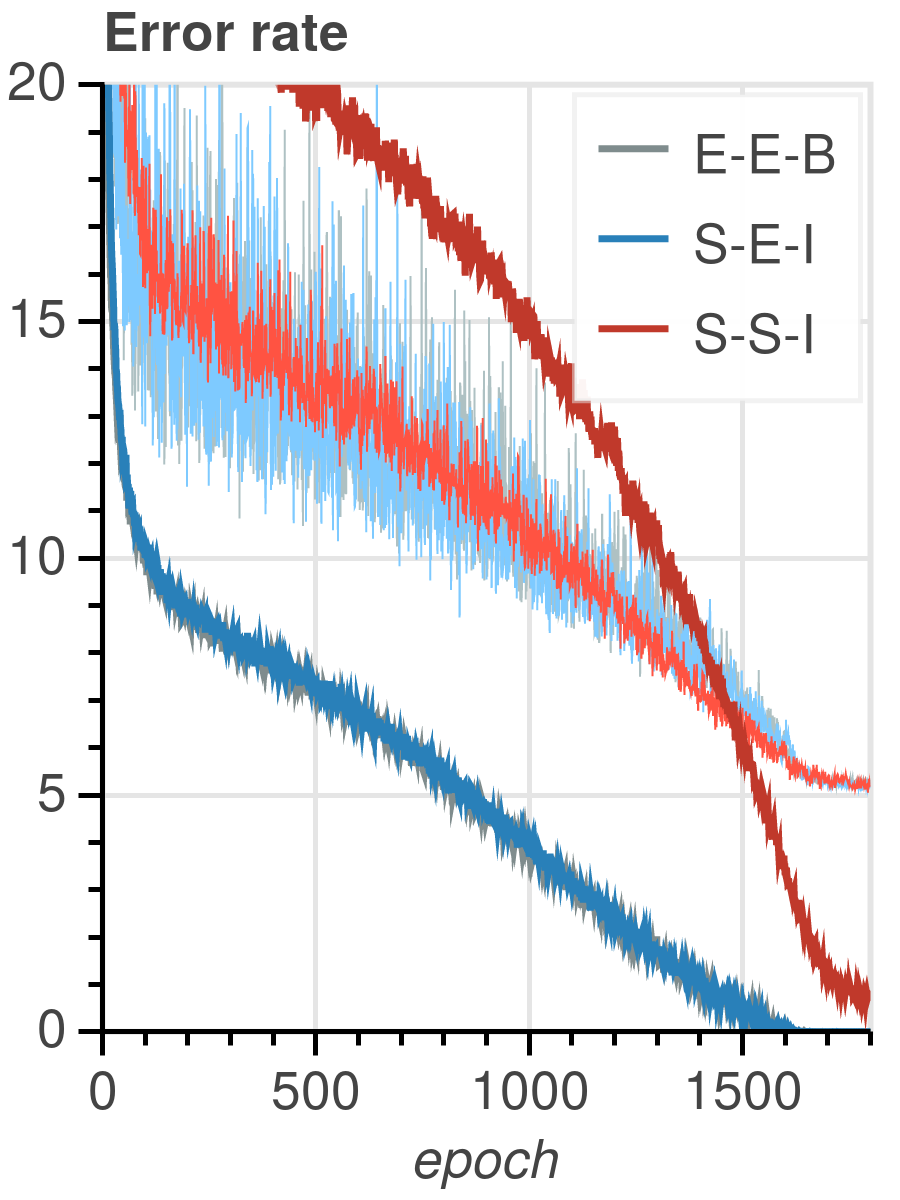}
    \end{subfigure}
    \begin{subfigure}[b]{0.32\textwidth}
        \includegraphics[width=\textwidth]{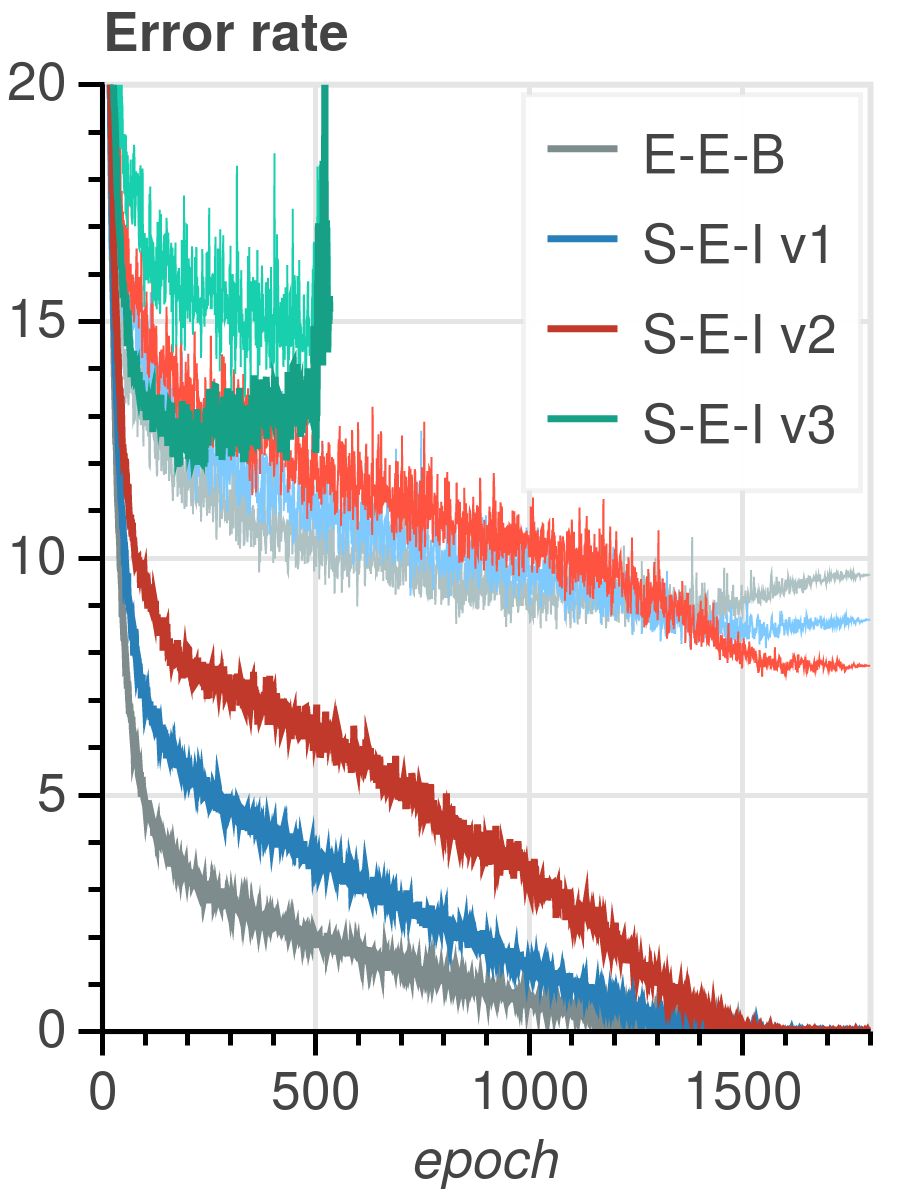}
    \end{subfigure}
    \caption{ Training curves (dark) and test curves (light). \textbf{Left:} Architecture A. \textbf{Center:} Architecture B. \textbf{Right:} Architecture C.}\label{fig:training_curves_noskip}
\end{figure}

Table~\ref{results5} and Figure~\ref{fig:training_curves_noskip} present the results of removing the skip connection. The first variant (A) is exactly like the 26 2x32d used on CIFAR-10 but without the skip connection (i.e. 2 branches with the following components {\tt ReLU-Conv3x3-BN-ReLU-Conv3x3-BN-Mul}). The second variant (B) is the same as A but with only 1 convolutional layer per branch ({\tt ReLU-Conv3x3-BN-Mul}) and twice the number of blocks. Models using architecture A were tested once and models using architecture B were tested twice.

\begin{table}
\caption{Error rates (\%) on CIFAR-10.}
\label{results5}
\begin{center}
\begin{tabular}{ccccc}
\toprule
\multicolumn{2}{c}{} &\multicolumn{3}{c}{Architecture} \\
\cmidrule(lr){3-5}
\multicolumn{1}{c}{Model} &\multicolumn{1}{c}{\(\alpha_{i.j}\)} &\multicolumn{1}{c}{A}  &\multicolumn{1}{c}{B} &\multicolumn{1}{c}{C}\\
\midrule
26 2x32d E-E-B	&n/a    &4.84	&5.17  &-\\
\cdashlinelr{1-5}
26 2x32d S-E-I	&rand(0,1)  &4.05	&5.09 &-\\
26 2x32d S-S-I	&rand(0,1)  &4.59 	&5.20 &-\\
\midrule
14 2x32d E-E-B	&n/a    &-	&-  &9.65\\
\cdashlinelr{1-5}
14 2x32d S-E-I v1	&rand(0.4,0.6)  &-	&- &8.7\\
14 2x32d S-E-I v2	&rand(0.35,0.65)  &- 	&- &7.73\\
14 2x32d S-E-I v3	&rand(0.30,0.70)  &- 	&- &diverges\\
\bottomrule
\end{tabular}
\end{center}
\end{table}

The results of architecture A clearly show that shake-shake regularization can work even without a skip connection. On that particular architecture and on a 26 2x32d model, S-S-I is too strong and the model underfits. The softer effect of S-E-I works better but this could change if the capacity is increased (e.g. 64d  or 96d).

The results of architecture B are actually the most surprising. The first point to notice is that the regularization no longer works. This, in itself, would indicate that the regularization happens thanks to the interaction between the 2 convolutions in each branch. The second point is that the train and test curves of the S-E-I and E-E-B models are absolutely identical. This would indicate that, for architecture B, the shake operation of the forward pass has no effect on the cost function. The third point is that even with a really different training curve, the test curve of the S-S-I model is nearly identical to the test curves of the E-E-B and S-E-I models (albeit with a smaller variance).

Finally, it would be interesting to see whether this method works without Batch Normalization. While batchnorm is commonly used on computer vision datasets, it is not necessarily the case for other types of problems (e.g. NLP, etc ..). Architecture C is the same as architecture A but without Batch Normalization (i.e. no skip, 2 branches with the following structure {\tt ReLU-Conv3x3-ReLU-Conv3x3-Mul}). To allow the E-E-B model to converge the depth was reduced from 26 to 14 and the initial learning rate was set to 0.05 after a warm start at 0.025 for 1 epoch. The absence of Batch Normalization makes the model a lot more sensitive and applying the same methods as before makes the model diverge. To soften the effect a S-E-I model was chosen and the interval covered by \(\alpha_{i.j}\) was reduced from [0,1] to [0.4,0.6]. Models using architecture C and different intervals were tested once on CIFAR-10. As shown in Table~\ref{results5} and Figure~\ref{fig:training_curves_noskip}, this method works quite well but it is also really easy to make the model diverge (see model 14 2x32d S-E-I v3).

\section{Conclusion}

A series of experiments seem to indicate an ability to combat overfit by decorrelating the branches of multi-branch networks. This method leads to state of the art results on CIFAR datasets and could potentially improve the accuracy of architectures that do not use ResNets or Batch Normalization. While these results are encouraging, questions remain on the exact dynamics at play. Understanding these dynamics could help expand the application field to a wider variety of complex architectures.

\newpage
\bibliography{shakeshake}
\bibliographystyle{shakeshake}

\end{document}